\documentclass[conference,letterpaper]{IEEEtran}

\usepackage{cite}
\usepackage[pdftex]{graphicx}
\graphicspath{{./figures/}}
\DeclareGraphicsExtensions{.pdf,.jpeg,.png}
\usepackage[cmex10]{amsmath}
\usepackage{array}
\usepackage{eqparbox}
\usepackage{stfloats}
\usepackage{float}
\usepackage[colorlinks = true,
            linkcolor = blue,
            urlcolor  = blue,
            anchorcolor = blue]{hyperref}
\usepackage{url}
\usepackage{graphicx}
\usepackage{subfig}
\usepackage{algorithm}
\usepackage{algorithmic}
\usepackage{setspace}
\usepackage{hyperref}
\usepackage{xcolor}
\usepackage{longtable}
\usepackage{enumerate}

\DeclareRobustCommand*{\IEEEauthorrefmark}[1]{\raisebox{0pt}[0pt][0pt]{\textsuperscript{\footnotesize #1}}}

\makeatletter
\newcommand*\titleheader[1]{\gdef\@titleheader{#1}}
\AtBeginDocument{%
  \let\st@red@title\@title
  \def\@title{%
    \bgroup\normalfont\large\centering\@titleheader\par\egroup
    \vskip0.5em\st@red@title}
}
\makeatother

\title{Operation-Level Performance Benchmarking of Graph Neural Networks for Scientific Applications}
\titleheader{Published as a workshop paper at MLSys 2022}

\begin{document}

\author{\IEEEauthorblockN{
Ryien Hosseini\IEEEauthorrefmark{1, 2}, 
Filippo Simini\IEEEauthorrefmark{1},  
Venkatram Vishwanath\IEEEauthorrefmark{1},   
}                                     

\IEEEauthorblockA{\IEEEauthorrefmark{1}
Argonne Leadership Computing Facility, Argonne National Laboratory, Lemont IL USA }

\IEEEauthorblockA{\IEEEauthorrefmark{2}
Department of Electrical Engineering and Computer Science, University of Michigan, Ann Arbor MI USA }

\newline \{rhosseini, fsimini, venkat\}@anl.gov

}

\maketitle

\thispagestyle{plain}
\pagestyle{plain}
 \begin{abstract}
 \boldmath
As Graph Neural Networks (GNNs) increase in popularity for scientific machine learning, their training and inference efficiency is becoming increasingly critical. Additionally, the deep learning field as a whole is trending towards wider and deeper networks, and ever increasing data sizes, to the point where hard hardware bottlenecks are often encountered. Emerging specialty hardware platforms provide an exciting solution to this problem. In this paper, we systematically profile and select low-level operations pertinent to GNNs for scientific computing implemented in the Pytorch Geometric software framework. These are then rigorously benchmarked on NVIDIA A100 GPUs for several various combinations of input values, including tensor sparsity. We then analyze these results for each operation. At a high level, we conclude that on NVIDIA systems: (1) confounding bottlenecks such as memory inefficiency often dominate runtime costs moreso than data sparsity alone, (2) native Pytorch operations are often as or more competitive than their Pytorch Geometric equivalents, especially at low to moderate levels of input data sparsity, and (3) many operations central to state-of-the-art GNN architectures have little to no optimization for sparsity. We hope that these results serve as a baseline for those developing these operations on specialized hardware and that our subsequent analysis helps to facilitate future software and hardware based optimizations of these operations and thus scalable GNN performance as a whole.    
 \end{abstract}

\IEEEpeerreviewmaketitle

\vspace{7pt}
\section{Introduction} \label{sec:intro}

In recent years, substantial research has been devoted to the development of Graph Neural Networks (GNNs) \cite{gnnsurvey}. GNNs have demonstrated to be powerful machine learning tools, which are able to learn robust representations of data in the form of graphs, and other graph-like data structures, such as point clouds, natural networks, and manifolds \cite{gnnsurvey,bronstein2017}. GNNs have applications in several diverse scientific tasks. For example, molecules, proteins, social networks, and knowledge networks can all be represented as graphs, and GNNs have been shown to accurately and efficiently solve many tasks in these applications \cite{gnnsurvey}, \cite{bronstein2017}. As the prominence of GNNs increase, so does the importance of its runtime performance and ability to extend such networks to larger data and model sizes. 

In this paper, we profile and analyze state-of-the-art GNN models used in several scientific applications, identify common operations of such models when implemented in Pytorch/Pytorch Geometric \cite{pytorch}, \cite{pygeo}, and then systemically benchmark these operations on a single NVIDIA A100 GPU. We evaluate operation benchmarks on several dimensions, such as input sizes and data sparsity. 
Thus, the primary purpose of this research is threefold:
\begin{enumerate}
    \item Aid software and hardware developers in identifying performance bottlenecks and thus focus effort on future improvements in GNN development tools,
    \item Provide machine learning scientists performance insights so as to optimize future GNN architectures
    \item Provide performance information on state-of-the-art hardware platforms for GNN inference.
\end{enumerate}

\subsection{Graph Neural Networks}
At a high level, graph neural networks extend common neural network operations, such as convolution, to non-euclidean data \cite{gnnsurvey}. While traditional deep learning applications such as computer vision operate on euclidean data, the generalization of common deep learning operations enable to use of these data-driven methods to solve a variety of scientific problems, in areas such as particle simulation, protein folding, traffic prediction, social-network related prediction tasks \cite{gnnsurvey}.

At the heart of GNNs is the message passing operation, wherein neighboring nodes of a graph are iteratively updated based on differentiable and permutation invariant functions \cite{gnnsurvey} \cite{pygeo}. Recently, a plethora of powerful GNN architectures have been developed. Indeed, many common deep learning architectures, such as recurrent networks, attention based methods, and variational autoencoders now have GNN equivalents. 

A principal advantage of GNNs is their relative computational efficiencies. Since the message passing algorithm operates on graphs, sparse inputs often require far less computation than their equivalent traditional DNNs on their euclidean-mapped counterparts. Thus, new operations that perform common operations (e.g. matrix multiplication, transpose, etc.) on sparsely formatted data have been introduced in many deep learning frameworks, such as Pytorch \cite{pytorch} \cite{pygeo}.

\subsection{Explicit Software Support for Sparse Operations}
We focus this work on the Pytorch \cite{pytorch} deep learning framework. Recently, as research interest in GNNs has increased, libraries such as Pytorch Geometric \cite{pygeo} have been developed in order to provide explicit support for creating GNNs and, by extension, sparse operations. Specifically, Pytorch Geometric is built directly on Pytorch and provides support for a diverse set of GNN development tasks, including end-to-end network implementations, efficient data handling operations, and optimizations of sparse operations that are often used by GNNs. 

Pytorch Geometric's operation level dependencies are  \texttt{torch\_sparse}, \texttt{torch\_scatter}, \texttt{torch\_cluster}, and \texttt{torch\_spline\_conv} libraries. These libraries  provide optimizations over existing Pytorch operations and contain newly implemented algorithms that are not available in standard Pytorch. 

In this paper, we make a distinction between so-called \textit{native} Pytorch operations and \textit{Pytorch Geometric} operations. Crucially, in cases where analogous operations exist between the two (e.g. \texttt{torch\_scatter.scatter\_add} and \texttt{torch.Tensor.scatter\_(op="add")}), we provide side-by-side performance benchmarks across several parameters such as data sparsity. We believe that the resulting analysis will be useful to understand the optimizations provided by the Pytorch Geometric library. 

It is important to note that, while outside the scope of this paper, alternatives to Pytorch Geometric exist, even within the Pytorch ecosystem. For instance, Deep Graph Library (DGL) \cite{dgl} provides native Python implementations of message passing kernels, which, perhaps at the performance costs of Pytorch/Pytorch Geometric's native C++ implementations, provide software platform agnostic support for message passing and other operations important to GNNs. In the future, the performance of other GNN libraries such as DGL can be compared to the existing Pytorch Geometric benchmarks found below. Additionally, we plan to expand scope to other popular deep learning frameworks such as TensorFlow. 

\subsection{Hardware Support for Sparse Operations}
A common problem facing many state-of-the-art deep learning systems (including GNNs), is how to efficiently train and deploy models with millions, or even billions, of learnable parameters \cite{ml_hpc}. As the size of neural networks continue to increase, so does the importance of efficient and scalable hardware systems to support them. 

Over the past several years, NVIDIA GPUs have become a de-facto standard hardware platform for training and deploying deep learning models at scale, due primarily to their hardware parallelism and support for common deep learning software tools, among other factors \cite{nvidia_and_dl_frameworks} \cite{nvidia_and_data_mining} \cite{nvidia_optim}. 

However, as models and their associated data continue to scale,  hardware bottlenecks have become increasingly problematic. In response, emerging specialized hardware compute platforms have attempted solve these problems of scalability and absolute performance. For example, the SambaNova RDU platform, Graphcore IPU architecture, and Cerebras CS-2 platform, have all attempted solve these efficiency bottlenecks by developing new dataflow architectures, expanding on device memory and compute power, and/or providing explicit hardware optimizations for deep learning \cite{sambanova_specific} \cite{sambanova} \cite{graphcore_perf} \cite{graphcore_applications} \cite{cerebras}. 
Therefore, rigorous operation-level performance benchmarks on NVIDIA systems can serve as baseline performance references in order to facilitate future development of such hardware systems.

\subsection{Profiling and Benchmark tools}
\label{sec:profile_and_bm_tools}

Previous work has used several profiling and benchmarking frameworks in order to evaluate both operations and applications (model) level performance of deep learning models. For example, in the domain of GNNs, \cite{dwivedibenchmarking} developed a generic benchmarking framework to facilitate applications level benchmarking of novel model architectures. Other work used more ad hoc methods, or used native Python tools such as the \texttt{time.timeit} module. 

In this paper, we elect to use the native Pytorch profiler and native Pytorch benchmark framework. This is due to several factors. Namely, since the scope of this paper is limited to Pytorch and Pytorch Geometric, we are not limited to generic profiling and benchmark tools. Additionally, the Pytorch profiler supports automatic operations level profiles of end-to-end networks, which is relevant to our native operation selection process (see \ref{sec:model_profiling} for more details). The Pytorch benchmark framework extends the native Python timing module to facilitate features such as automatic warmup runs, thread synchronisation, and other details critical to equitable and accurate benchmarking. 

However, future work focusing on software frameworks other than Pytorch will require a generic benchmark framework. Please refer to \ref{sec:future_work} for more details.

\vspace{7pt}
\section{Related Work} \label{sec:lit}
In this section, we discuss prior work in profiling GNNs and benchmarking low-level operations relevant to GNNs and other deep neural networks. 

\subsection{Benchmarking Sparse Operations}
Previous work has focused on benchmarking sparse operations at the kernel level, benchmarking of specific novel sparse operations, and hardware optimizations for sparse operators. For example, \cite{sparse_kernel_bm} demonstrates the relative performance increases of NVIDIA's cuSPARSE backend compared to the standard cuBLAS, as well as proposes further kernel optimizations for certain operations. Additionally, \cite{newspspmm} proposed new software optimizations to improve hardware efficiency of a particular operation, sparse-sparse matrix multiplication. Other groups, such as \cite{sparse_hardware}, propose several hardware architecture improvements to accelerate sparse operations, such as by unifying memory management to support continguously represented sparse data. Mainstream vendors such as NVIDIA have also proposed hardware improvements to their systems in order facilitate more efficient sparse operations. However, to our knowledge, no group has performed systematic benchmarks for low-level operations critical to GNN performance at the systems level and within a specific deep learning framework.

\subsection{GNN Benchmarking}
Since the rise of popularity of GNNs, substantial research has focused on creating effective and equitable benchmark paradigms. For instance, \cite{dwivedibenchmarking} created one of the first generic and open source benchmark frameworks, wherein researchers can easily add novel models and datasets and rigorously compare results over end-to-end models, datasets, and hardware systems. Frameworks such as these are critical for the development of more efficient GNN architectures. However, such frameworks are generally concerned with end-to-end model performance and stop short of providing operations level benchmarks for GNNs.

Other previous work has focused on creating comprehensive and equitable datasets in order to facilitate equitable end-to-end model performance. Notably, \cite{ogb} introduced \textit{Open Graph Benchmark}, a set of diverse graph benchmarks that can be used to assess model performance across several metrics. These datasets are especially valuable in that they include both realistic synthetic data that may be used for stress testing new models and hardware systems as well as real world data that may be used for downstream model performance analysis.

% op level stuff -- pytorchgeo paper benchmarks, but not much as far as rigorous ops level benchmarking yet
Perhaps most relevant to our work is the intial benchmarking conducted by \cite{pygeo} , wherein Pytorch geometric implementations of several GNN architectures were compared to existing implimentations, such as those of Deep Graph Library (DGL) \cite{dgl}. Additionally, \cite{pygeo}  provided benchmark operations-level comparisons of their scatter and gather operations. While they stopped short of comprehensive operations level benchmarks, the results they did provide enables us to robustly compare and resemblance check our results.

\vspace{7pt}
\section{Experimental Design}

The benchmarking pipeline of this paper can be broken into two main sections. First, common graph neural network (GNN) models are \textit{profiled} in order to determine key operations for GNNs (and sparse operations in general). We use these results in order to collate these operations along with otheroperations that are specifically designed to facilitate efficient sparse operations in PyTorch Geometric. 

Second, we systematically benchmark each of these operations on NVIDIA A100 GPUs. In the following subsections, we motivate and detail the procedure for each of these steps.

\subsection{Model Profiling}
\label{sec:model_profiling}
The first motivation is to profile low-level operations that are integral to modern GNN models. Thus, we select and robustly benchmark several end-to-end GNN architectures. Relevant literature was reviewed in order to determine a set of 8 models that we believe provide a prudent representation of the current state-of-the-art in the field. Specifically, representations of different scientific applications (molecules, social networks, traffic), graph prediction task (node classification, node regression, graph regression, link prediction, etc.), and architecture (graph attention, VAEs, spectral and spatial methods) were considered. Additionally, models were screened by their perceived impact on the field (e.g. citation count, use in industry, etc.) as well as ease of implementation in the Pytorch Geometric framework. The final models selected for profiling can be found in table \ref{tab:models_selected}.

\begin{table}[]
\centering
\caption{Selected GNN models for operations-level profiling}
\label{tab:models_selected}
\scalebox{0.66}{
\begin{tabular}{|l|l|}
\hline
\textbf{Model}                                                       & \textbf{Primary Domain} \\ \hline
Graph Attention Network (GATv2) \cite{gatv2}                             & Proteins   \\ \hline
Temporal Graph Network (TGN) \cite{tgn}                                & Social Networks   \\ \hline
Principal Neighbourhood Aggregation (PNA) \cite{pna} & Vision   \\ \hline
Dynamic Neighborhood Aggregation  (DNAConv) \cite{dnaconv}     & Several (representation learning)   \\ \hline
Graph U-Net \cite{graphunet}                                                & Several (representation learning)  \\ \hline
SchNet \cite{schnet}                                                      & Quantum Interactions   \\ \hline
Directional Message Passing Network (DimeNet) \cite{dimenet}                   & Molecules   \\ \hline

\end{tabular}
}
\end{table}

Each model was then profiled using the native Pytorch profiler, as discussed in \ref{sec:profile_and_bm_tools}. Each model was run on synthetic data of varying sparsity, first with default model parameters. Then, the models were extended such that they utilized full GPU memory. Specifically, models were extended by increasing network parameters (iteratively deepening and widening each network). Where extending networks lead to negligible performance increases, data batch sizes were increased so as to reach full GPU utilization. Top operations were then collated for each model and were used to select the native Pytorch operations for benchmarking (\ref{sec:native_op_selection}). Results from profiling these models are detailed in \ref{sec:results_a}.

\subsection{Operation Selection}
\label{sec:op_selection}
In this section, we detail the process used to select relevant sparse operations. As previously noted, Pytorch Geometric has four dependencies which contain optimized operations for GNNs: \texttt{torch\_sparse}, \texttt{torch\_scatter}, \texttt{torch\_cluster}, and \texttt{torch\_spline\_conv}. In order to select operations for benchmarking, we examined each of these libraries and selected fundamental operations that were considered low-level and pertinent to GNNs. Crucially, any available native Pytorch implementations of these operations were also benchmarked in order to facilitate comparison between the two implementations. Note that the \texttt{torch\_cluster} library, which contains efficient implementations of many high-level cluster operations, is not covered in this initial benchmark experiment. 

Additionally, the Pytorch operations which contributed the most to the overall compute time across all models during label profiling (\ref{sec:model_profiling}) were evaluated for criteria such as availability of sparse implementations for the CUDA backend, relevance to other sparse operators, overlap with the operations selected above, etc. Based on these constraints, five native Pytorch operations were also selected for benchmarking.

The list of all operations selected for benchmarking can be found in \ref{sec:selected_ops} and \ref{sec:native_op_selection}.

\subsection{Operation Benchmarking}
\label{sec:op_bm}

In this section, we describe the experimental setup for operation benchmarking. As discussed in \ref{sec:lit}, experimental details of operations-level benchmarking are rarely standardized, and to our knowledge, no standard practice for this task exists. Thus, throughout this section, we justify design decisions in-so-far as possible. 

We test each operation at the \textit{full memory capacity of the machine} in order to facilitate fair comparison of operations across different hardware platforms. We note that typical use cases of these operations may often use inputs far smaller than those tested here. However, given the goal of many deep learning accelerator hardware systems are to \textit{extend} common GNN architectures, benchmarking at GPU capacity facilitates comparison to hardware accelerator systems that often have far higher memory capacity than standard GPUs.
We define this as $\geq 95\%$ memory capacity. Thus, for each pair of operation and dimension (e.g. one dimensional tensors, two dimensional tensors, etc), we select the input size so that the combination of input values with the highest memory utilizes $\geq 95\%$ of the total memory capacity of the machine. All other (less expensive) combinations of input values with the same dimension for that operation will retain the same input size in order to allow for robust comparisons of different operation parameters.

It is important to note that this later point implies that many of the input combinations may yield peak memory loads of less than 95\%. For example, benchmarking a fully dense tensor will often require substantially more memory overhead than a comparatively sparser one (even in cases where the they are represented as strided tensors). However, in experiments where we benchmark an operation over several sparsity values, we fix the input tensor to be of size such that the \textit{most dense} tensor utilizes full memory capacity. Thus, more sparse tensors of the same size will \textit{not} utilize full capacity. This enables clear comparison of the effect of sparsity on the operation runtime. 

After fixing the input size for different tensor shapes and varying tensor sparsity, we also, where relevant, vary the other parameters of the operation. However, in some cases, not all parameters are tested, especially in cases where either relevant literature or our operation profiling indicated typical use cases for an operation. We note that while more comprehensive data is always beneficial, it is often not feasible to test and analyze all parameters of every operation due to the polynomial increase in benchmark points required for each new parameter. 

Thus, the high-level process to determine the inputs to all operations is as follows:
\begin{enumerate}
    \item Identify parameters to vary for the specific operation, always including sparsity of the input tensor. Other parameters depend on the specific operation (e.g. dimensionality of tensor, reduce operation, etc.).
    \item Given these combinations of input parameters, identify the combination (for all input shapes) with highest peak memory capacity. 
    \item Increase size of input tensors until this input combination reaches full memory capacity ($\geq 95\%$).
    \item Set input tensor size for all other input combinations to the value above.
\end{enumerate}

There are a number of additional assumptions and details associated with our benchmarking method, such as data types, hardware details, etc. that are summarized below:

\begin{enumerate}
    \item Pytorch version: \texttt{1.9.0}
    \item CUDA version: \texttt{11.3}
    \item Pytorch Geometric version: \texttt{2.0.4}
    \item NVIDIA GPU details: NVIDIA A100-SXM4-40GB (40.536 DDR). 
    \item All operations where benchmarked at least \textbf{14} times and all reported value are the median value benchmarked.
    \item All operations benchmarked have an interquantile range of $\leq 0.00017$ seconds across single benchmark iterations. 
    \item All tensors are assumed to be of type \texttt{torch.float32} (data) or \texttt{torch.int64} (indices) unless otherwise noted.
\end{enumerate}

Additionally, we note that for this paper, we only benchmark the \textit{forward pass} of each operation, and do not include values for the \textit{backwards pass}.

Finally, we define the term \textit{reduce factor (RF)} to refer to the combination of inputs needed in order for the ratio of the total elements of the \textit{input} tensor to the \textit{output} tensor of an operation to be approximately the reduce factor. For example, a reduce factor of 8 indicates that the output tensor of an operation is about 8 times smaller than its input.

\section{Results: Profiling and Operation Selection} 
\label{sec:results_a}

\subsection{Operation Selection}

Below, we provide lists of the selected operations based on the method described in the previous section.

\subsubsection{Sparse Operations}
\label{sec:selected_ops}
The final operations selected for benchmarking from the \texttt{torch\_sparse}, \texttt{torch\_scatter}, and \texttt{torch\_spline\_conv} libraries can be found in Table \ref{tab:sparse_ops_selected} below.
\begin{table}[h]
\centering
\caption{Selected Sparse Operations for Benchmarking}
\label{tab:sparse_ops_selected}
\scalebox{0.55}{
\begin{tabular}{|l|l|l|l|}
\hline
Operation     & Library             & Native Pytorch Counterpart                 & Notes                    \\ \hline
scatter\_max  & torch\_scatter      &                                            &                          \\ \hline
scatter\_min  & torch\_scatter      &                                            &                          \\ \hline
scatter\_mean & torch\_scatter      &                                            &                          \\ \hline
scatter\_add  & torch\_scatter      & torch.Tensor.scatter\_ (reduce="add")      & Native impl. Inplace     \\ \hline
              &                     & torch.Tensor.scatter\_ (reduce="multiply") & Native impl. Inplace     \\ \hline
coalesce      & torch\_sparse       & torch.Tensor.coalesce                      &                          \\ \hline
transpose     & torch\_sparse       & torch.Tensor.transpose                     & with COO formatted input \\ \hline
spmm          & torch\_sparse       & torch.smm                                  &                          \\ \hline
spspmm        & torch\_sparse       & torch.sparse.mm                            &                          \\ \hline
spline\_conv  & torch\_spline\_conv &                                            &                          \\ \hline
\end{tabular}
}
\end{table}

\subsubsection{Selected \texttt{Pytorch}/\texttt{Aten} Operations}
\label{sec:native_op_selection}
The final operations selected for benchmarking from Pytorch can be found in Table \textbf{\ref{tab:native_ops_selected}} below.

\begin{table}[h]
\centering
\caption{Selected Native Pytorch Operations for Benchmarking}
\label{tab:native_ops_selected}
\scalebox{0.65}{
\begin{tabular}{|l|l|l|}
\hline
Torch Level Operation Name & Analagous aten:: Operation & Notes                     \\ \hline
torch.addmm                & aten::addmm                &                           \\ \hline
torch.index\_select        & aten::index\_select        &                           \\ \hline
torch.sort                 & aten::sort                 &                           \\ \hline
torch.Tensor.index\_add\_  & aten::index\_add\_         & In-place tensor operation \\ \hline
torch.gather               & aten::gather               &                           \\ \hline
\end{tabular}
}
\end{table}

\section{Results: Operation Benchmarking}
\label{sec:results_b}
In the following subsections, the results of benchmarking the operations listed in \ref{sec:selected_ops} and \ref{sec:native_op_selection} are shown and analyzed. For each operation, or group of operations, trends in data that were found to be particularly insightful are shown and discussed. However, full benchmark numbers are omitted for brevity and can instead be found at our project repository site \footnote{\href{https://github.com/ryienh/gnn-ops-benchmark}{https://github.com/ryienh/gnn-ops-benchmark}}. Additionally, visualizations for some operations are omitted but can also be viewed in the project repository site. 

\subsection{Selected Native Pytorch Operations}
\label{sec:native_results}

This section explores the Pytorch operations selected as part of model profiling in \ref{sec:results_a}. These operations were found during profiling to be commonly used in end-to-end GNN architectures but are not part of Pytorch Geometric or its associated dependencies.

\subsubsection{\texttt{torch.sort}} \label{op_result:sort}
Sorting operations have long been a focus of algorithms and systems/performance research. The Pytorch sort operation sorts a single strided input tensor. The documentation and source code for this operation can be found \href{https://pytorch.org/docs/stable/generated/torch.sort.html}{here}. 
Figure \ref{fig:native_sort} provides a summary of the operation profile. Our benchmarking experiments yield the following principal outcomes:

\begin{figure}[t]
    \centering
    \includegraphics[type=pdf,ext=.pdf,read=.pdf,scale = .5]{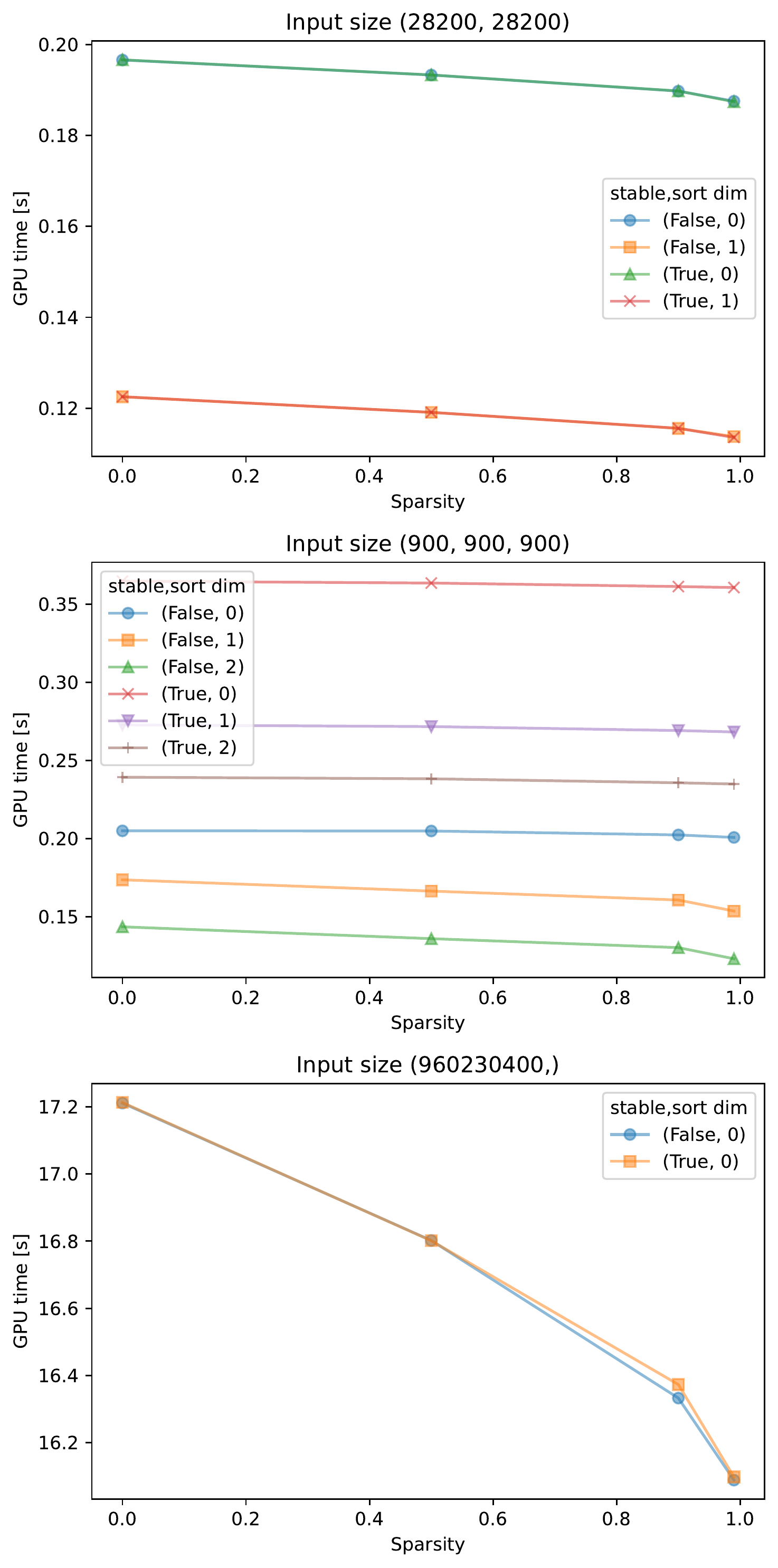}
    \caption{
    Benchmarking results for the native \texttt{torch.sort} operation. This operation is benchmarked for input dimensionalities between 1 and 3 and for input sparsity values of 0, 0.5, 0.9, and 0.99. Additionally, the implementations for stable and unstable sort are explicitly compared. 
    \label{fig:native_sort}}
    \end{figure}

\begin{enumerate}[(i)]
    \item In general, comparatively more sparse input tensors result in faster execution times. This is probably due to the fact that sparser tensors have larger chunks of zero elements that are already sorted. 
    \item The larger the dimensionality of the input tensor, the less the above effect is observed, with the inverse correlation between input tensor sparsity and operation runtime diminishing substantially. We hypothesize that this is due to more intensive swap and copy operations, which bottleneck the algorithm and thus reduce the significance of the input sparsity on the total runtime. 
    \item As expected, the stable sorting algorithm results in slower execution times compared to its unstable counterpart.
    \item As only densely formatted tensors are supported for this operation, the sparsity of the input does not affect system memory. 
    \item Sorting across higher dimensions of the input result in comparatively faster execution times. This is consistent with the C/C++ style memory hierarchy. However, in order to fully reconcile this phenomenon with the CUDA memory hierarchy, further analysis focusing on information such as cache misses, memory organization, etc. is necessary. Please see \ref{sec:benchmarkdetail} for more details. 
\end{enumerate}
Given the observations above, we believe that the primary bottlenecks that should be targeted for future operation optimizations are related to memory overheads. The implementation of a sort algorithm that operates on sparsely formatted tensors will substantially improve memory and runtime overheads, especially in multidimensional cases. 

\subsubsection{\texttt{torch.addmm}}
\begin{figure}[h!]
    \centering
    \includegraphics[type=pdf,ext=.pdf,read=.pdf,scale = .5]{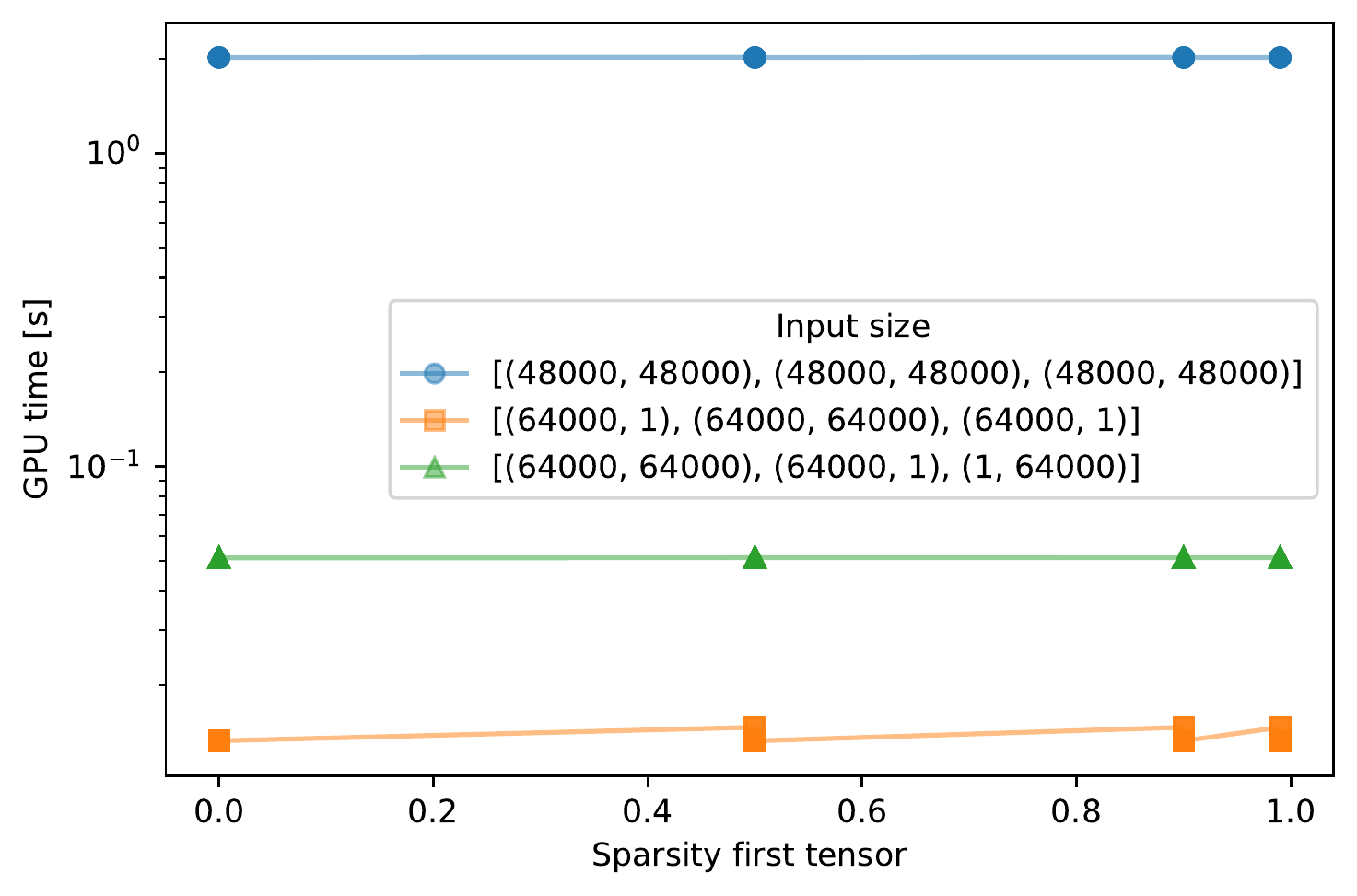}
    \caption{
    Benchmarking results for the native \texttt{torch.addmm} operation. This operation is benchmarked for different combinations of input dimensionalities between 1 and 2 and for input sparsity values of 0, 0.5, 0.9, and 0.99.
    \label{fig:native_addmm}}
    \end{figure}

This operation combines a matrix multiplication with an add operation. Specifically, the operation matrix multiplies two strided input tensors and adds the result to a third strided input tensor. Thus, our benchmarking tests different input combinations of 1 and 2 dimensions (using broadcasting where applicable). The documentation and source code for this operation can be found \href{https://pytorch.org/docs/stable/generated/torch.addmm.html}{here}.  Figure \ref{fig:native_addmm} provides a summary of the operation profile. Our benchmarking experiments yield the following principal outcomes:

\begin{enumerate}[(i)]
    \item As all tensors are strided tensors, sparsity has no discernible effect on runtime or memory performance.
    \item As expected based on theoretical runtime analysis, performing matrix multiplication on higher dimensional tensors result in substantially larger runtimes. Varying the add tensor results in a far smaller, yet non-negligible runtime increase.
\end{enumerate}
The current multiply-and-add operation does not optimize for sparse tensor input. Thus, implementation of a version of this operation for sparsely formatted tensors will likely substantially improve memory overhead and runtime performance for sparse tensor inputs. 

\subsubsection{\texttt{torch.gather}}
\label{op_result:gather}

At a high level, this operation gathers values of an input tensor by indexing according to a given index tensor input. Our benchmarking test includes different input combinations of 1, 2 and 3 dimensions and, in the latter two cases, varies the dimension which is indexed. The documentation and source code for this operation can be found  \href{https://pytorch.org/docs/stable/generated/torch.gather.html}{here}. Our benchmarking experiments yield the following principal outcomes:

\begin{enumerate}[(i)]
    \item Given the same total memory capacity, operations across a single dimension result in comparatively larger runtimes vs. multidimensional tensors. This is due to the fact that the absolute number of gather operations is larger when gathering across a single dimension. 
    \item The above result extends to the total dimensionality of the input tensor more generally. Specifically, when controlling for all other variables, the number of dimensions of the input tensor is inversely proportional to the runtime. This is a direct consequence of (i). 
    \item When gathering across a higher dimension, runtimes are comparatively longer. We hypothesize that this is due to GPU memory storage hierarchy and access factors. However, additional analysis in future work is required in order to explore this further (See \ref{sec:benchmarkdetail}).
    \item The index dimensionality parameter dominates the input dimensionality parameter. Thus, low index dimensions with low dimensional tensor inputs outperform high index dims with high dimensional tensor inputs.
    \item Due to strided tensor input formats, input sparsity has negligible effect on total runtime and memory overhead.
\end{enumerate}
Similar to other native Pytorch operations, we again observe negligible effects of input tensor sparsity and runtime. Additionally, this operation currently only supports strided tensor inputs, thus fixing the storage capacity of the tensor regardless of input sparsity. Consequently, we believe future implementation of a gather operation for sparsely formatted tensors will greatly improve performance of this operation when used with sparse inputs.  

\subsubsection{\texttt{torch.index\_select}}
\label{op_result:idx_select}

This operation uses an inputted \textit{index} vector in order to index an inputted value tensor along a given dimension. Our benchmarking test includes different input combinations of 1, 2 and 3 dimensions and, in the latter two cases, varies the dimension which is indexed. Additionally, we test several \textit{reduce factor (RF)} values, as defined in \ref{sec:op_bm}. The documentation and source code for this operation can be found \href{https://pytorch.org/docs/stable/generated/torch.index_select.html}{here}. Our benchmarking experiments yield the following principal outcomes:

\begin{enumerate}[(i)]
    \item Higher reduce factors lead to faster runtimes. This is due to the fact that the number copy operations decreases linearly with the size of the output. Additionally, the total operation memory overhead also decreases due to the decreased output size of the operations at high reduce factor values.
    \item Indexing at higher dimensions results in higher runtimes. Similar to \ref{op_result:gather}, this is likely due to memory hierarchy details of the GPU system, which is outside the scope of this paper but should be explored further in future work. 
    \item Unlike, \ref{op_result:gather}, reduce factor dominates index dimension. That is, high index dimensions with high reduce factors outperform low index dimensions with low reduce factors. This can be explained by comparing the indexing of the \texttt{torch.gather} and \texttt{torch.index\_select} operations: The former indexes across an arbitrary number of index dimensions while the later uses the (1-dimensional) index tensor to index the value tensor. Thus, the memory access order is reversed.
    \item Due to strided tensor input formats, sparsity has negligible effect on total runtime and memory overhead.
\end{enumerate}
We again observe negligible effects of input tensor sparsity and runtime due to lack of support for sparsely formatted tensors and no explicit software optimization for sparse tensors with strided format. Again, we believe the implementation of a \texttt{index\_select} operation for sparsely formatted tensors will greatly improve performance of this operation when used with sparse inputs.  

\subsubsection{\texttt{torch.Tensor.index\_add\_}}
At a high level, this operation adds the value of an inputted \textit{source} tensor to the \texttt{self} tensor according to the indices given in the \textit{index} input. Our benchmarking test includes different input combinations of 1, 2 and 3 dimensions and, in the latter two cases, varies the dimension which is indexed. The documentation and source code for this operation can be found \href{https://pytorch.org/docs/stable/generated/torch.Tensor.index_add_.html}{here}. The associated visualizations for this operation are excluded from this document for brevity but can be referenced in our project repository. Our benchmarking experiments yield the following principal outcomes:

\begin{enumerate}[(i)]
    \item Indexing at higher dimensions results in higher runtimes. This is consistent with the memory hierarchy discussion outlined in \ref{op_result:gather} and \ref{op_result:idx_select}.
    \item There is no discernible pattern between dimensionality of input and the runtime. This is a counterintuitive result that warrants future investigation in future work. However, one hypothesis may be that discrepancies for total element size of the input tensors may cause this issue. Specifically, the memory overhead required for the multi-dimensional cases is such that there is substantial $\geq 10\%$ discrepancies in total element size. Thus, future work should normalize for total element size as well as investigate the memory sub-system profiles further here.
    \item Due to strided tensor input formats, sparsity has negligible effect on total runtime and memory overhead.
\end{enumerate}
As seen in the previous native operations, we again observe negligible effects of input tensor sparsity and runtime due to lack of support for sparsely formatted tensors and no explicit software optimization for sparse tensors with strided format. Again, we believe the implementation of a \texttt{index\_select} operation for sparsely formatted tensors will greatly improve performance of this operation when used with sparse inputs. Additionally, we encounter the counterintuitve result wherein dimensionality of input has no correlation with runtime performance. Future research is necessary in order to analyze and verify this claim. 

\textit{High-level Conclusions}: The profiled native Pytorch operations only support strided tensor inputs. With the exception of the sort operation (\ref{op_result:sort}), sparsity of the input tensors had negligible effects on runtime performance. Given that these operations were shown in \ref{sec:model_profiling} to be a prominent part of state-of-the-art GNNs, implementation of sparse equivalents of these operations have the potential to improve end-to-end GNN performance, especially when the input to these operations are sparse. 

\subsection{Scatter operations} \label{sec:scatter_results}
This section explores the benchmark results of scatter operations in the \texttt{torch\_scatter} package, one of the four dependencies of the Pytorch geometric package. Scatter operations are a type of reduce operation that map an input tensor to an output tensor by using an \textit{index tensor input}. Where duplicate indices exist, a \textit{reduce} operation is used to select the final output value at that index.

All scatter operations are benchmarked by varying the following parameters: input dimension (1 and 2 dimensions), input sparsity, and \textit{reduce factor} (as defined in \ref{sec:op_bm}). Where available, we provide direct comparisons to the analogous native Pytorch operation, \texttt{torch.Tensor.scatter\_}. We also note that unlike the \texttt{torch\_scatter} operations, \texttt{torch.Tensor.scatter\_} operations are \textit{in-place}.

The documentation and source code for these operations can be found at the following hyperlinks: \href{https://pytorch.org/docs/stable/generated/torch.scatter.html}{\texttt{torch\_scatter}}  and \href{https://pytorch.org/docs/stable/generated/torch.Tensor.scatter_.html}{\texttt{torch.Tensor.scatter\_}}.

\subsubsection{\texttt{scatter\_min} and \texttt{scatter\_max}}
\begin{figure}[t]
    \centering
    \includegraphics[type=pdf,ext=.pdf,read=.pdf,scale = .5]{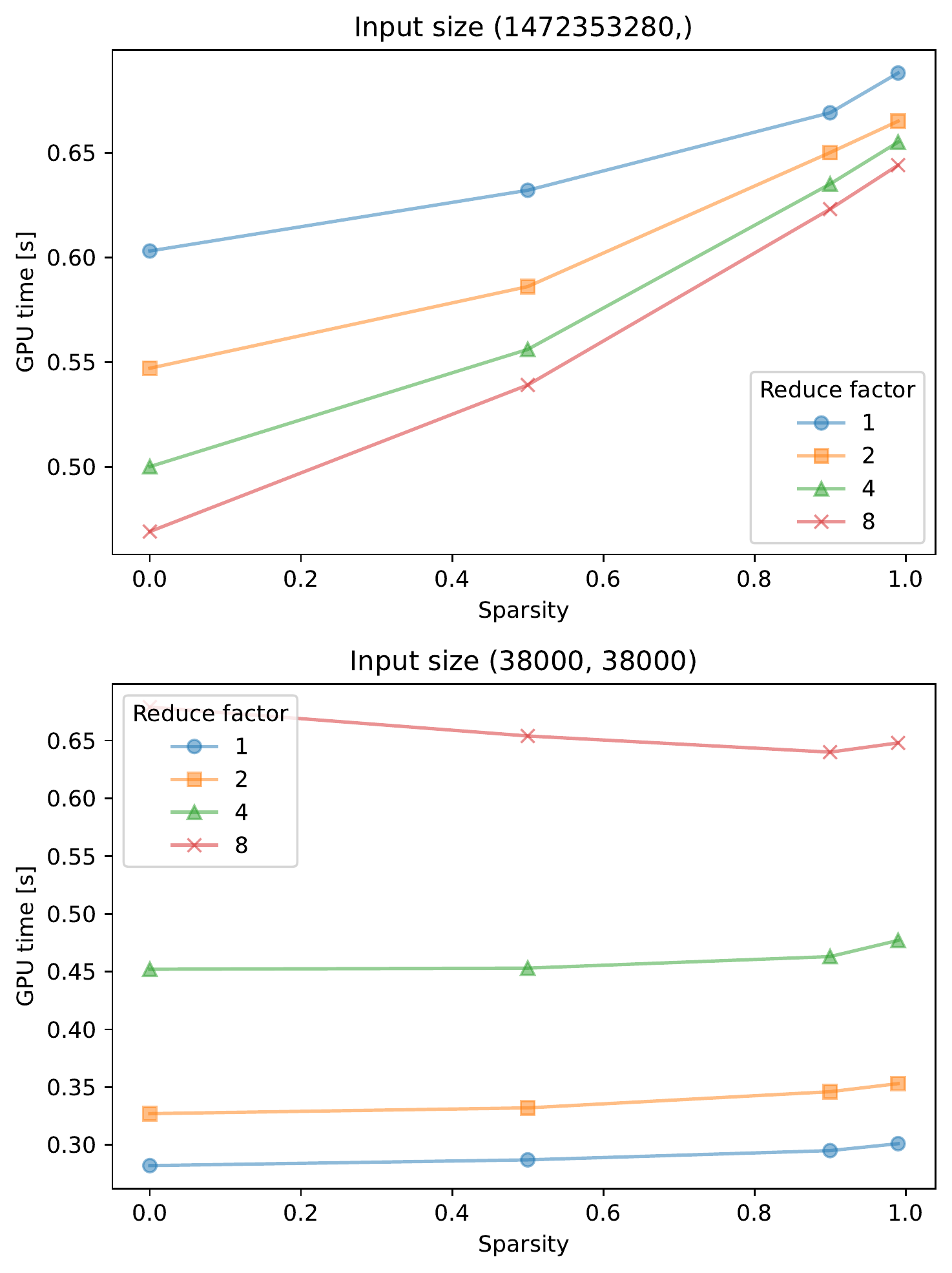}
    \caption{
    Benchmarking results for the \texttt{scatter\_min} operation. This operation is benchmarked for inputs of both 1 and 2 dimensions, reduce factors of 1, 2, 4, and 8, and input sparsity values of 0, 0.5, 0.9, and 0.99.
    \label{fig:scatter_min}}
    \end{figure}

We first evaluate the results of benchmarking the \texttt{scatter\_min} and \texttt{scatter\_max} operations. The \texttt{scatter\_min} is visualized in Figure \ref{fig:scatter_min}. 

\begin{enumerate}[(i)]
    \item In the one-dimensional case, higher reduce factors lead to comparatively \textit{faster} runtimes and smaller total operation memory overhead. Similar to \ref{op_result:idx_select}, this is due to the fact that the number copy operations decreases linearly with the size of the output.
    \item In the two dimensional case, higher reduce factors lead to comparatively \textit{slower} runtimes. This is may be due to the fact that the reduce operations are calculated across two dimensions and are thus non-contiguous. However, additional examination of the GPU memory system is necessary in order to confirm this hypothesis. 
    \item Runtime and sparsity have a direct relationship, which is unexpected. 
   Change in runtime with respect to sparsity is most pronounced in low dimensional cases. This may be due to the fact that higher dimensional tensors require far more copy operations per step (one must copy entire tensor at $dim \neq idx\_dim$ to output). Thus, this bottlenecks the algorithm. 
    \item As expected, \texttt{scatter\_min} and \texttt{scatter\_max} have similar trends in runtime and memory peaks due to the reduce operations being nearly identical in computation time.
\end{enumerate}
The observations above imply that sparse inputs lead to comparatively \textit{inferior} runtime speeds, perhaps due to memory non-contiguity and irregular accesses. Additionally, the effect of the reduce factor on runtime varies depending on the dimensionality of the input. Thus, we believe that \textit{memory hierarchy} optimizations may be the most promising course of action to improving performance the performance of these operations in end-to-end GNN models. 

\subsubsection{\texttt{scatter\_mean} and \texttt{scatter\_add}}
\begin{figure}[t]
    \centering
    \includegraphics[type=pdf,ext=.pdf,read=.pdf,scale = .5]{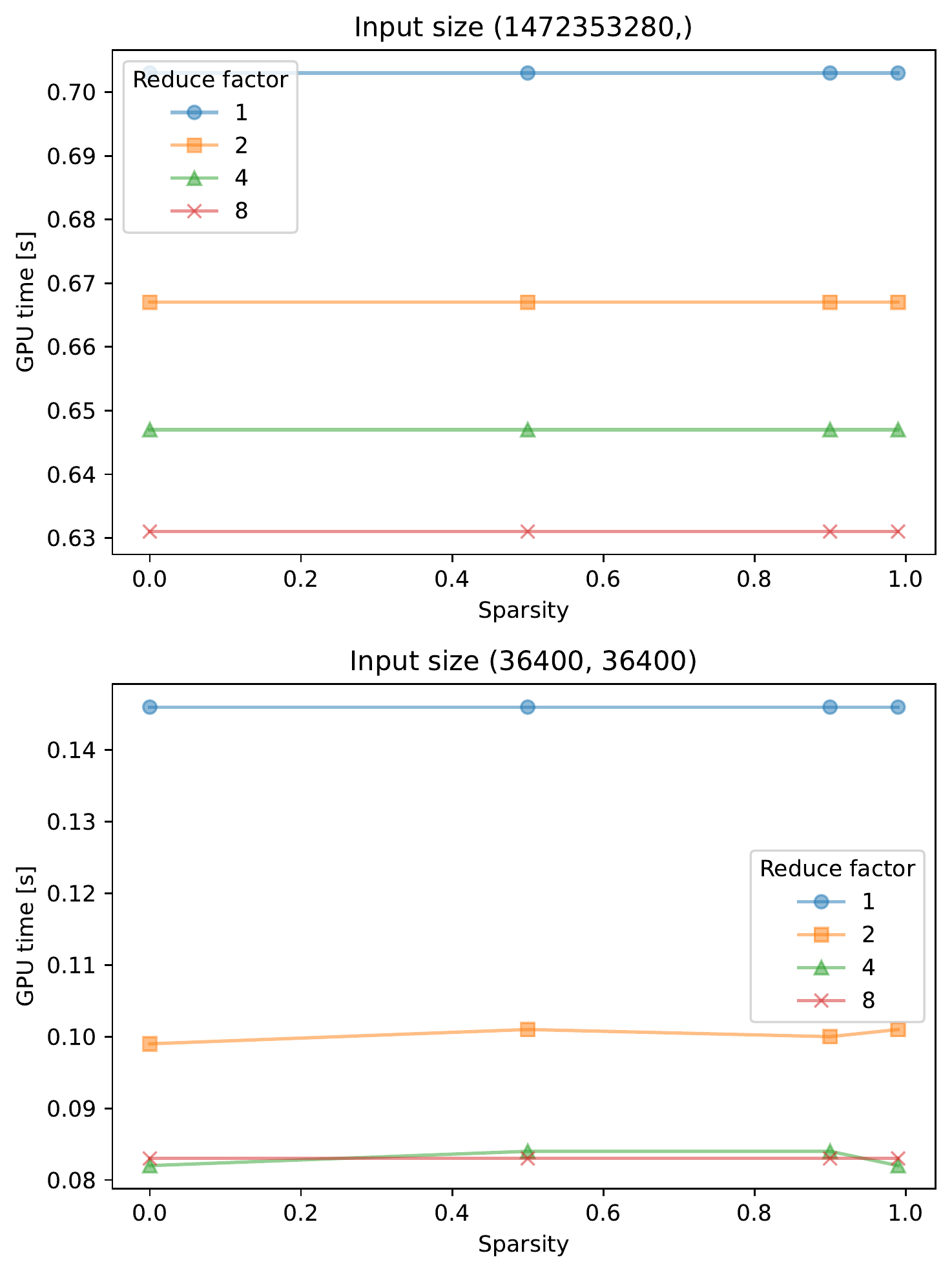}
    \caption{
    Benchmarking results for the \texttt{scatter\_mean} operation. This operation is benchmarked for inputs of both 1 and 2 dimensions, reduce factors of 1, 2, 4, and 8, and input sparsity values of 0, 0.5, 0.9, and 0.99.
    \label{fig:scatter_mean}}
    \end{figure}

We now evaluate the results of benchmarking the \texttt{scatter\_mean} and \texttt{scatter\_add} operations. Additionally, \texttt{scatter\_mean} is visualized in Figure \ref{fig:scatter_mean}. We note that the analogous native Pytorch implementation to \texttt{scatter\_add} is \texttt{torch.Tensor.scatter\_ (reduce="add")}, and both operations' results are included in our analysis. 

\begin{enumerate}[(i)]
    \item Both \texttt{scatter\_mean} and \texttt{scatter\_add} have negligible changes in runtime as input sparsity is varied. We hypothesize that this is due to mean and add operations requiring more computation compared to min and max. Additionally, analysis of the source code indicates that optimization for 0 valued elements is present in the latter but not the former. 
    \item We also compare \texttt{scatter\_add} with the native Pytorch operation \texttt{torch.Tensor.scatter\_ (reduce="add")}. Performance of the native operation is comparatively faster than than \texttt{scatter\_add}. This is likely largely due to the in-place nature of the native operation.
\end{enumerate}

Unlike with the min and max scatter operations, varying the sparsity of the input has only negligible effects on overall runtime. Thus, we believe the biggest computational bottlenecks for these operations are either the reduce operations themselves or the memory hierarchy of the GPU. 

\subsubsection{\texttt{torch.Tensor.scatter\_ (reduce="multiply")}}
We also benchmark the native \texttt{torch.Tensor.scatter\_ (reduce="multiply")}, which has no \texttt{torch\_scatter} analogue. We note that as with the earlier add and mean scatter operations, varying the sparsity of \texttt{torch.Tensor.scatter\_ (reduce="multiply")} has a negligible effect on the total runtime and that the effect of the reduce factor depends on the dimensionality of the input. 

\textit{High-level Conclusions} The chosen reduction operation (e.g. min, max, add, etc.) of benchmarked scatter operations significantly influence the effect of other factors, such as sparsity and input dimensionality of the input, on the overall runtime. Examination of source code indicates significantly different implementation decisions based on the reduction operation selected. Thus, simplification to a single efficient implementation may improve performance of many of these operations. 

Additionally, as described in the above sections, significant variance in runtime is observed across different input dimensionalities (despite approximately constant total number of elements and reduce operations required). Thus, these operations may benefit from memory hierarchy designs that are developed for sparse inputs. 

\subsection{Sparse operations}
\label{sec:sparse_results}
This section explores the benchmark results of miscellaneous sparse operations in the \texttt{torch\_sparse} package, one of the dependencies of the Pytorch geometric package. Specifically, we examine sparse-dense and sparse-sparse matrix multiplication, sparse matrix transpose, and sparse coalesce operations. All of the operations in this section contain analogous native Pytorch operations, which are included in the analysis. We note that all native Pytorch operations use sparse (COO) formatted tensors.

\subsubsection{\texttt{sparse\_spmm} and \texttt{sparse\_spspmm}}

\begin{figure}[t]
    \centering
    \includegraphics[type=pdf,ext=.pdf,read=.pdf,scale = .5]{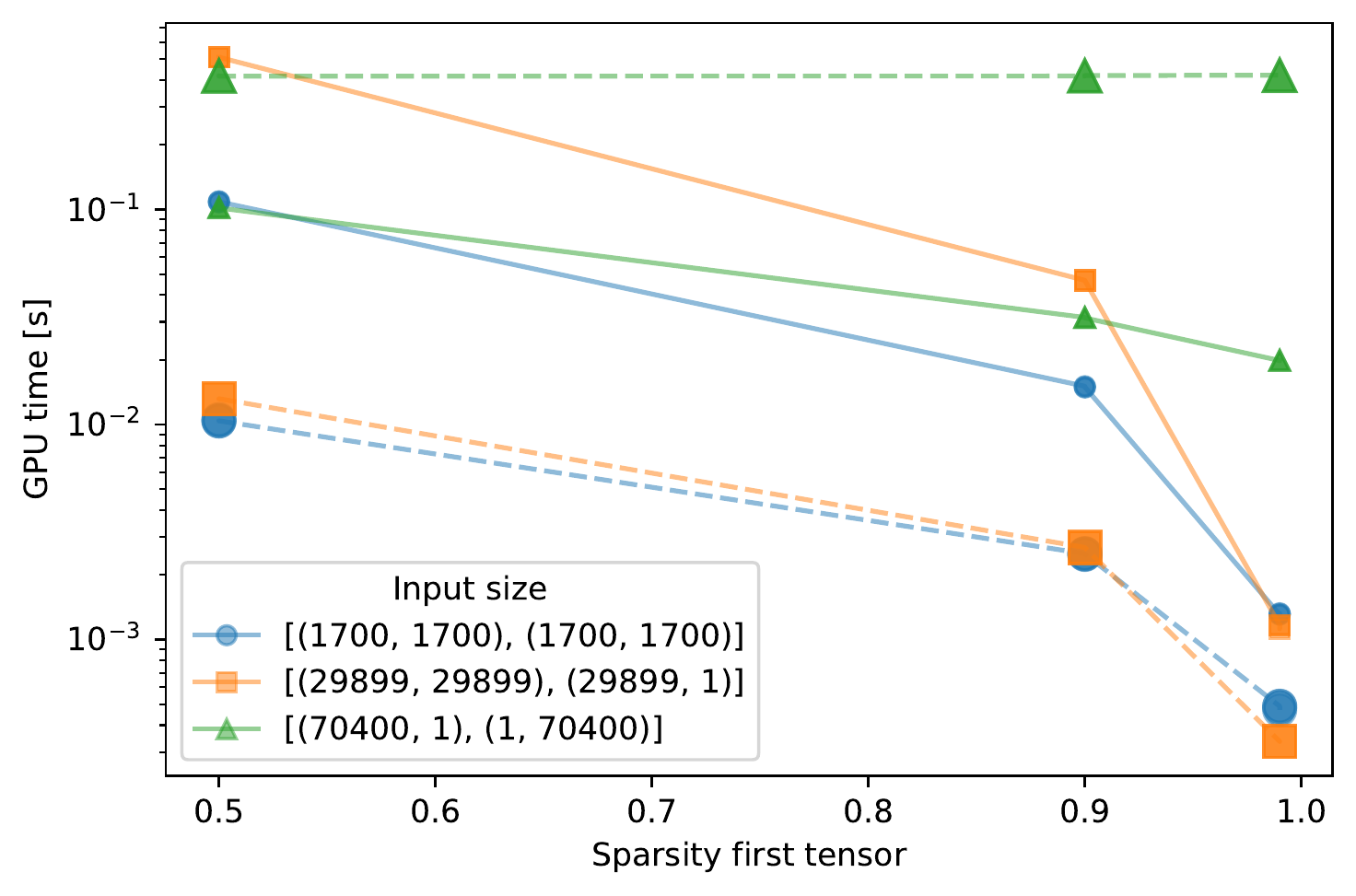}
    \caption{
    Benchmarking results for the \texttt{sparse\_spmm} operation. This operation is benchmarked for combinations of inputs of 1 and 2 dimensions and input sparsity values of 0, 0.5, 0.9, and 0.99. Solid lines and small markers denote PyTorch Geometric operations, whereas dashed lines and large markers denote PyTorch operations. 
    \label{fig:sparse_spmm}}
    \end{figure}

\begin{figure}[t]
    \centering
    \includegraphics[type=pdf,ext=.pdf,read=.pdf,scale = .5]{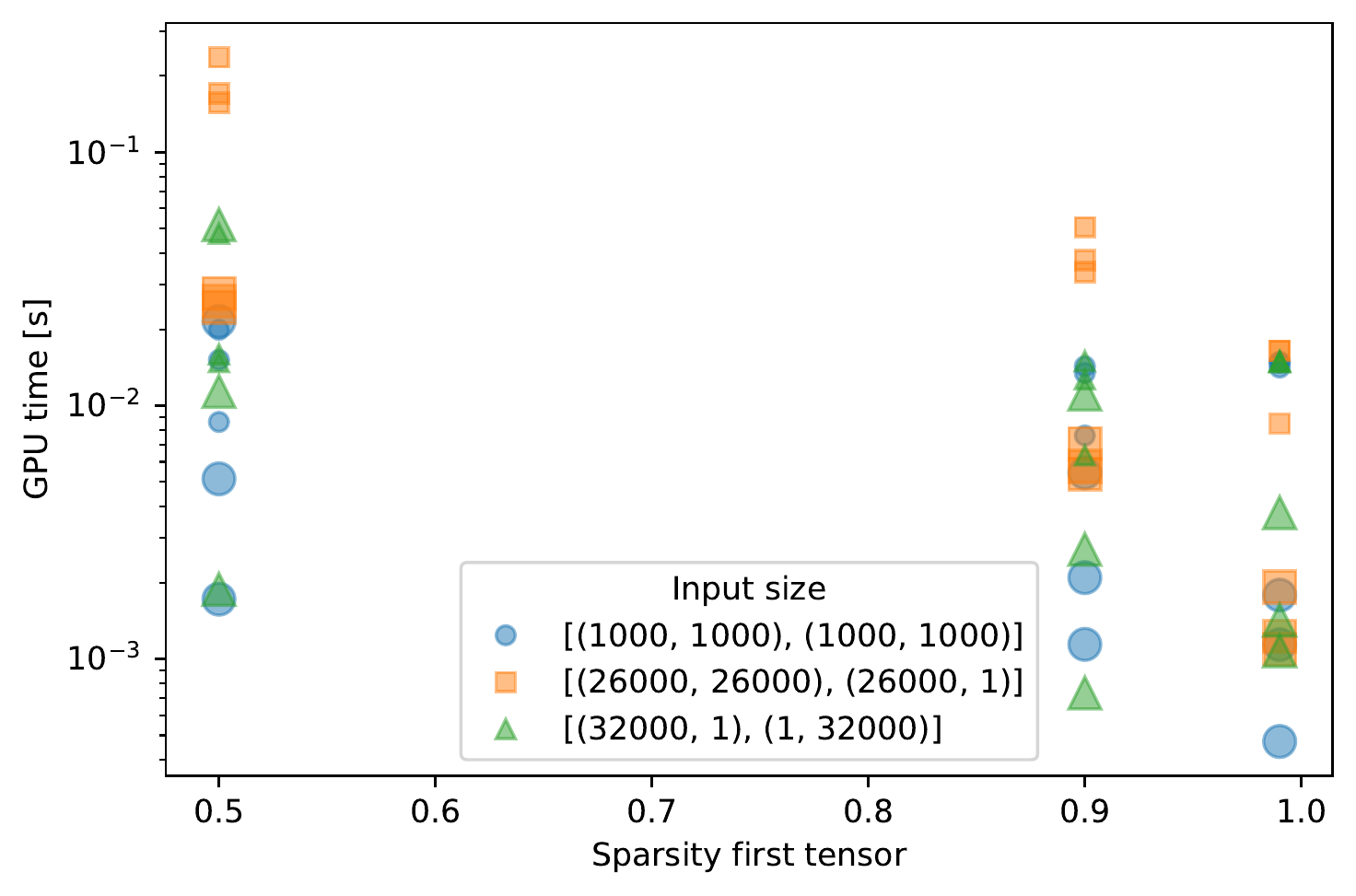}
    \caption{
    Benchmarking results for the \texttt{sparse\_spspmm} operation.  This operation is benchmarked for combinations of inputs of 1 and 2 dimensions and input sparsity values of 0, 0.5, 0.9, and 0.99. Small markers denote PyTorch Geometric operations, large markers denote PyTorch operations. 
    \label{fig:sparse_spspmm}}
    \end{figure}

We first evaluate the results of benchmarking the \texttt{sparse\_spmm} and \texttt{sparse\_spspmm} operations, which correspond to sparse-dense and sparse-sparse matrix multiplication respectively. Both operations are analogous to \texttt{torch.sparse.mm} (as this operation can operate on both COO and strided tensor inputs). Thus, we compare \texttt{sparse\_spmm} to \texttt{torch.sparse.mm} with a single COO and single strided tensor input and \texttt{sparse\_spspmm} to \texttt{torch.sparse.mm} with two COO formatted inputs. These results are visualized in Figures \ref{fig:sparse_spmm} and \ref{fig:sparse_spspmm}, respectively.  The documentation and source code for these operations can be found \href{https://github.com/rusty1s/pytorch_sparse}{here}, and for the native operation \href{https://pytorch.org/docs/stable/generated/torch.sparse.mm.html}{here}.

\begin{enumerate}[(i)]
    \item \texttt{sparse\_spmm} underperforms its native equivalent in all but the 1 dimensional input case. 
    \item In general, \texttt{sparse\_spspmm} also underperforms its native equivalent, across all sparsity levels.
    \item The sparsity of the first tensor (for sparse-dense multiplication) and the overall sparsity (for sparse-sparse multiplication) is inversely proportional to both runtime and memory requirements. 
    \item This effect is most pronounced at high sparsity levels (e.g. between 90\% and 99\% sparsity). Thus, future work can investigate performance differences across sparsity levels close to 100\%.
\end{enumerate}

The above results imply that the native Pytorch \texttt{torch.sparse.mm} operation often outperforms its Pytorch Geometric counterpart. Thus, the runtime performance of end-to-end GNN models on GPUs will likely improve with either further optimization of the Pytorch Geometric operations or use of the native Pytorch operations.  

\subsubsection{\texttt{sparse\_transpose}}
We now evaluate the sparse matrix transpose operation \texttt{sparse\_transpose} and its native Pytorch analogue \texttt{torch.transpose}. The documentation and source code for this operation and its analogous Pytorch native equivalent can be found \href{https://github.com/rusty1s/pytorch_sparse}{here} and  \href{https://pytorch.org/docs/stable/generated/torch.transpose.html}{here}, respectively. We highlight the following observations:

\begin{enumerate}[(i)]
    \item The Pytorch Geometric transpose operation under-performs its native counterpart. This effect is more pronounced at comparatively lower input sparsity values. 
    \item For both operations, runtime increases logarithmically with the input sparsity. 
    \item The Pytorch Geometric operation runtime performance suffers significantly in the two-dimensional input case. For example, for a two-dimensional input size and sparsity of 50\%, the Pytorch Geometric operation is on the order of 200 times slower than its native Pytorch counterpart. We hypothesize that this may be due to lack of memory optimization for copy operations for the Pytorch Geometric operation. 
\end{enumerate}

Based on these results, we conclude that, as with the sparse matrix multiplication operations, that the use of native Pytorch operations will likely improve end-to-end GNN performance. 

\subsubsection{\texttt{sparse\_coalesce}}
Next, we evaluate the coalesce operation \texttt{sparse\_coalesce} and its native Pytorch analogue \texttt{torch.coalesce}. These results are visualized in Figure \ref{fig:sparse_coalesce}. The documentation and source code for this operation and its analogous Pytorch native equivalent can be found \href{https://github.com/rusty1s/pytorch_sparse}{here} and  \href{https://pytorch.org/docs/stable/generated/torch.Tensor.coalesce.html}{here}, respectively. We highlight the following observations:

\begin{figure}[t]
    \centering
    \includegraphics[type=pdf,ext=.pdf,read=.pdf,scale = .5]{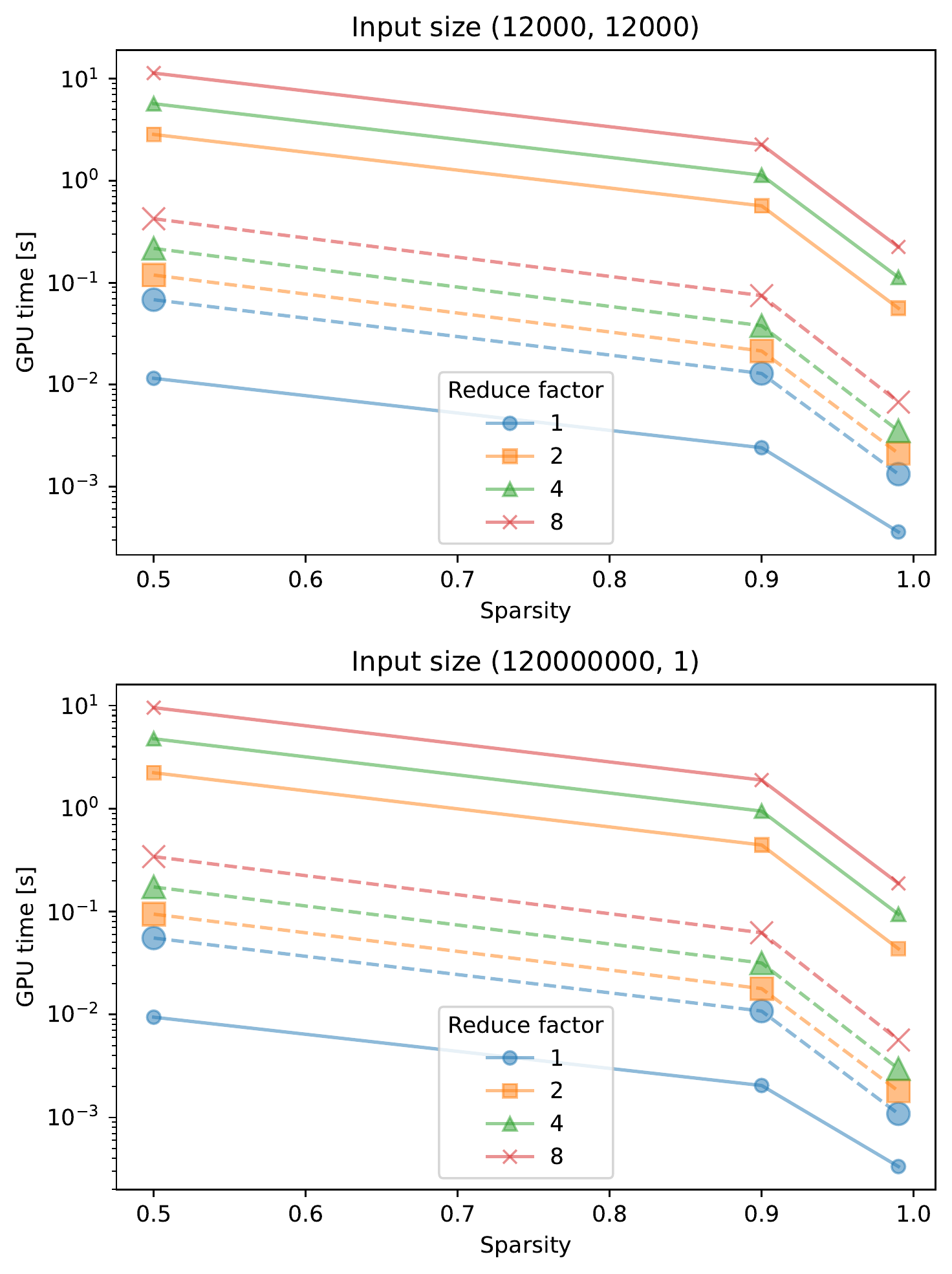}
    \caption{
    Benchmarking results for the \texttt{sparse\_coalesce} operation.  This operation is benchmarked for inputs of 1 and 2 dimensions, reduce factors of 1, 2, 4, and 8, and input sparsity values of 0, 0.5, 0.9, and 0.99. Solid lines and small markers denote PyTorch Geometric operations, whereas dashed lines and large markers denote PyTorch operations. 
    \label{fig:sparse_coalesce}}
    \end{figure}

\begin{enumerate}[(i)]
    \item For both operations, runtime decreases with the input sparsity. This effect is especially pronounced between 90\% and 99\% input sparsity levels. 
    \item For all input shapes, a higher reduce factor leads to comparatively higher runtimes. This is due to the fact that a higher reduce factor requires a comparatively higher number of add operations in order to coalesce the tensor. 
    \item For all but the trivial case where the reduce factor is 1, the native Pytorch Geometric operation under-performs its native Pytorch equivalent. This result holds across all input tensor sparsity levels.
\end{enumerate}
We again conclude that the native Pytorch implementation is competitive with the Pytorch Geometric implementation on GPU and should be considered in order to improve GNN runtime performance. 

\textit{High-level Conclusions}: Based on the \texttt{torch\_sparse} operations benchmarking, we note that, unlike in previous operation groups, a key bottleneck is input tensor sparsity. Thus, further software and hardware optimization should focus on efficiency of especially sparse (and specifically COO formatted) tensors. Additionally, we find that for all operations benchmarked, native Pytorch operations are either competitive with, or outperform, the Pytorch Geometric operations. Thus, the use of these operations should be considered as part of a larger GNN implementation and future work should focus on optimization of the Pytorch Geometric operations. 

\subsection{Spline Convolution Operation}
This section focuses on benchmarking the Pytorch Geometric implementation of the \textit{spline convolution operator}. The spline convolution operator is a generalization of deep neural networks that operates on graph structured data, and, by utilizing B-splines, is able to decouple kernel size from computation time \cite{spline_conv}. The associated Pytorch Geometric operation is \texttt{torch\_spline\_conv.spline\_conv} and its documentation can be found \href{https://github.com/rusty1s/pytorch_spline_conv}{here}.

We benchmark this operator over input sparsity and kernel size parameters. We note that, unlike the preceding operations, sparsity is presented in the form of \textit{average node degree} which is \textit{inversely proportional} to sparsity. Additionally, due to differences in total node size in our stochastically generated synthetic dataset, we present our results as GPU time normalized by the total node size of the input. Our key observations can be found below:
\begin{figure}[t]
    \centering
    \includegraphics[type=pdf,ext=.pdf,read=.pdf,scale = .5]{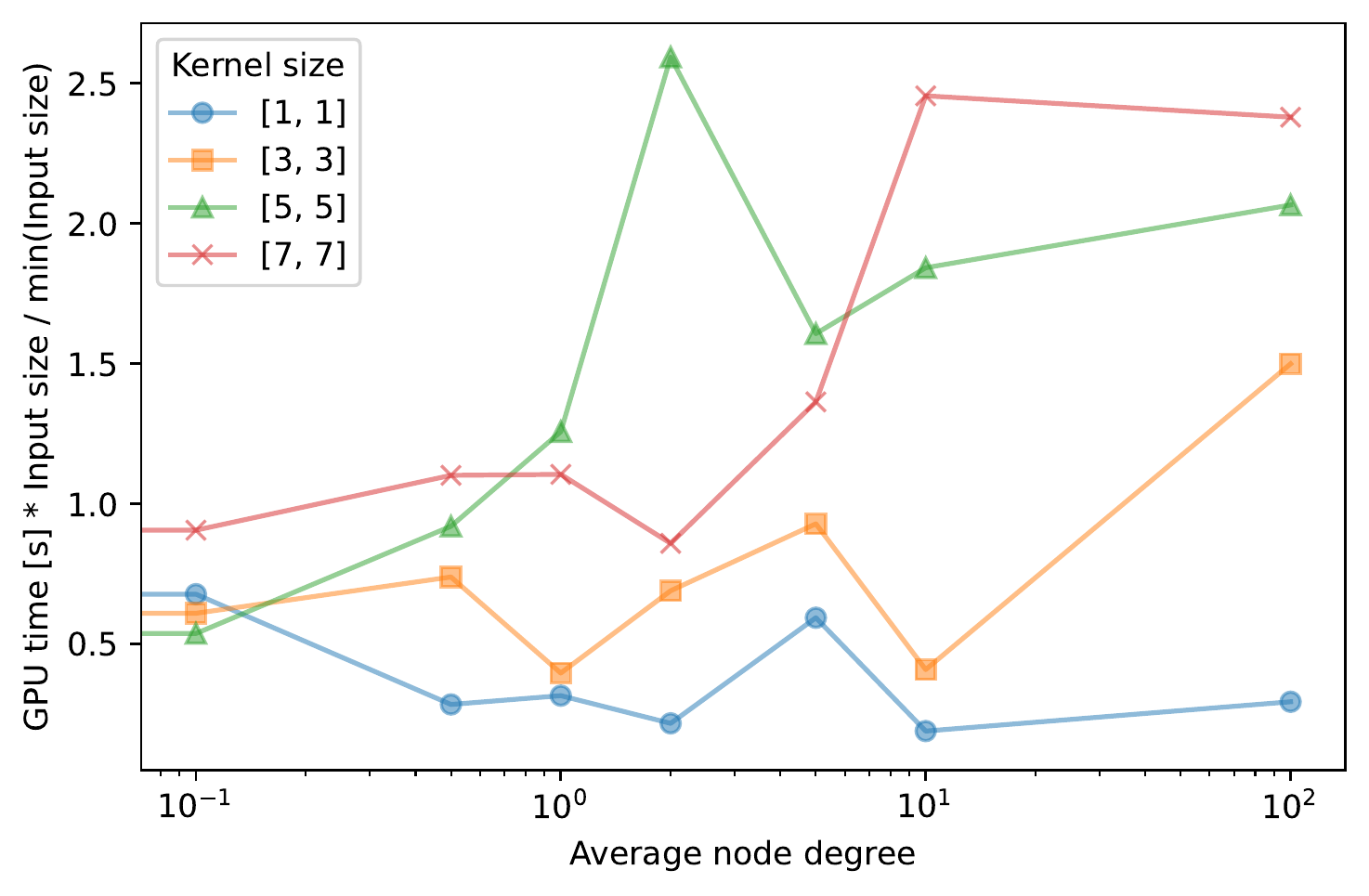}
    \caption{
    Benchmarking results for the \texttt{spline\_conv} operation.  This operation is benchmarked for kernel sizes of 1, 3, 5 and 7, and input sparsity values of 0, 0.5, 0.9, and 0.99.
    \label{fig:spline_conv}}
    \end{figure}

\begin{enumerate}[(i)]

    \item Total runtime is directly proportional to the average node degree and thus inversely proportional to the input sparsity. 
    \item Larger convolution kernels result in comparatively longer absolute runtimes, especially at higher levels of input data sparsity. 
\end{enumerate}

While the above observations are fairly intuitive, we hope that the absolute benchmark numbers provided will help facilitate further hardware and software based optimization of this operation.

\vspace{7pt}
\section{Future Work}
\label{sec:future_work}
While we believe that the operation profiling and benchmark framework proposed in this paper, along with the results and associated analysis and insights, is a valuable contribution, it is not comprehensive over all operations relevant to GNNs. Rather, we believe it lays the groundwork for future study of operations level performance across different software frameworks, hardware platforms, model architectures, etc. In this section, we outline areas of improvement for future work in the area. 

\subsection{Operation Benchmarks in Training Regime}

This paper focuses solely on the forward pass of each operation during benchmarking. Thus, performance is limited to the \textit{inference} stage of GNNs. Future work should build on our framework in order to facilitate benchmarking of the backwards pass implementations of each operation as well. We believe this should be a relatively simple extension of the current implementation of the benchmarking framework. 

\subsection{Expanding Operation Coverage}
Future work can expand on the results and analysis found here for other sparse operations. For example, the \texttt{torch\_cluster} library and other top operations profiled in \ref{sec:model_profiling} may serve as starting points for further research in this direction. 

\subsection{Additional Benchmark Metrics}
\label{sec:benchmarkdetail}
Throughout this paper, several hypotheses were proposed to explain empirical results of each operation. Thorough investigation of these hypotheses will require benchmarking with additional systems metrics, such as cache misses, data representation in memory, and specific hardware bottlenecks in each operation. 

\subsection{Extending Comparisons to Other Hardware Platforms}
We hope that the GPU results described in this paper serve as baseline performances for those developing these operations on specialized hardware. Additionally, our analysis may also prove valuable to those developing optimized hardware architectures to support sparse operations. Future work may focus on benchmarking these operations on existing hardware platforms in order to evaluate their performances compared to a single NVIDIA A100 machine.

\vspace{7pt}
\section*{Conclusion}
\label{sec:conclusion}
In conclusion, we provide the following high-level analysis of our operation benchmarks, summarizing the operation level observations of section (\ref{sec:results_b}):
\begin{enumerate}
    \item In several cases, such as in \ref{sec:native_results} and \ref{sec:scatter_results}, memory hierarchy order dominates sparsity. Thus, in these cases, bottlenecks identified for optimization should perhaps focus on memory, rather than solely input data sparsity. 
    \item Native Pytorch operations should be considered competitive, often matching or outperforming their Pytorch Geometric counterparts. This is especially true when data is not very sparse, and holds even when the native Pytorch equivalent is not an in-place operation. This performance gap is especially pronounced for sparse operations such as transpose and sparse multiply, where native operations support COO formatted sparse tensor representations.
    \item While some ops with strided inputs are optimized for comparatively sparser inputs (e.g. \texttt{torch.sort}), many (such \texttt{torch.addmm}) have no such optimizations. This can cause both memory and runtime bottlenecks which can significantly affect end-to-end GNN model performance. Thus, GNNs will likely benefit from sparse implementations of the benchmarked native operations.
\end{enumerate}
We plan to expand our efforts to cover additional accelerator architectures and model architectures in the future.

\section*{Acknowledgements}
This research used resources of the Argonne Leadership Computing Facility, which is a DOE Office of Science User Facility supported under Contract DE-AC02-06CH11357.

\bibliographystyle{unsrt} 
\bibliography{refs}

\end{document}